\documentclass{article}

\usepackage{microtype}
\usepackage{graphicx}
\usepackage{subfigure}
\usepackage{booktabs} %
\usepackage{enumitem}

\usepackage{hyperref}

\usepackage[accepted]{icml2025}

\usepackage{amsmath}
\usepackage{amssymb}
\usepackage{mathtools}
\usepackage{amsthm}
\usepackage{tikz}
\usepackage{hyperref}
\usepackage[utf8]{inputenc}
\usepackage{filecontents}
\usepackage{csvsimple}
\usepackage{pgfplots} %
\usepackage{xspace}
\usetikzlibrary{shapes}

\usepackage[capitalize,noabbrev]{cleveref}

\newcommand{\molmoe}{\textsc{Mol-MoE}\xspace}

\usepackage[textsize=tiny]{todonotes}

\begin{document}

\icmltitlerunning{Training Preference-Guided Routers for Molecule Generation}

\twocolumn[
\icmltitle{Mol-MoE: Training Preference-Guided Routers for Molecule Generation}

\begin{icmlauthorlist}
    \icmlauthor{Diego Calanzone}{udem,mila}
    \icmlauthor{Pierluca D'Oro}{mila}
    \icmlauthor{Pierre-Luc Bacon}{udem,mila}
\end{icmlauthorlist}

\icmlaffiliation{mila}{Mila Quebec AI Institute}
\icmlaffiliation{udem}{Université de Montréal}

\icmlcorrespondingauthor{Diego Calanzone}{diego.calanzone@mila.quebec}

\icmlkeywords{Deep Learning, Computational Biology, Drug Discovery, Mixture of Experts, Multi-objective optimization}

\vskip 0.3in
]

\printAffiliationsAndNotice{} %

\begin{abstract}

Recent advances in language models have enabled framing molecule generation as sequence modeling.
However, existing approaches often rely on single-objective reinforcement learning, limiting their applicability to real-world drug design, where multiple competing properties must be optimized. Traditional multi-objective reinforcement learning (MORL) methods require costly retraining for each new objective combination, making rapid exploration of trade-offs impractical.
To overcome these limitations, we introduce \molmoe, a mixture-of-experts (MoE) architecture that enables efficient test-time steering of molecule generation without retraining. Central to our approach is a preference-based router training objective that incentivizes the router to combine experts in a way that aligns with user-specified trade-offs. This provides improved flexibility in exploring the chemical property space at test time, facilitating rapid trade-off exploration. Benchmarking against state-of-the-art methods, we show that \molmoe achieves superior sample quality and steerability.%

\end{abstract}

\section{Introduction}

De-novo drug design is inherently a multi-objective optimization problem \citep{KernsDi2008}. While drug discovery often focuses on pharmacological efficacy, successful therapeutics must simultaneously optimize across multiple axes: selectivity, clinical safety, and toxicity profiles. These compounds must exist within a highly constrained subset of chemical space that satisfies all these criteria, making the search for viable drug candidates particularly challenging and computationally intensive.

\begin{figure}[t]
    \centering
    \includegraphics[width=0.9\linewidth]{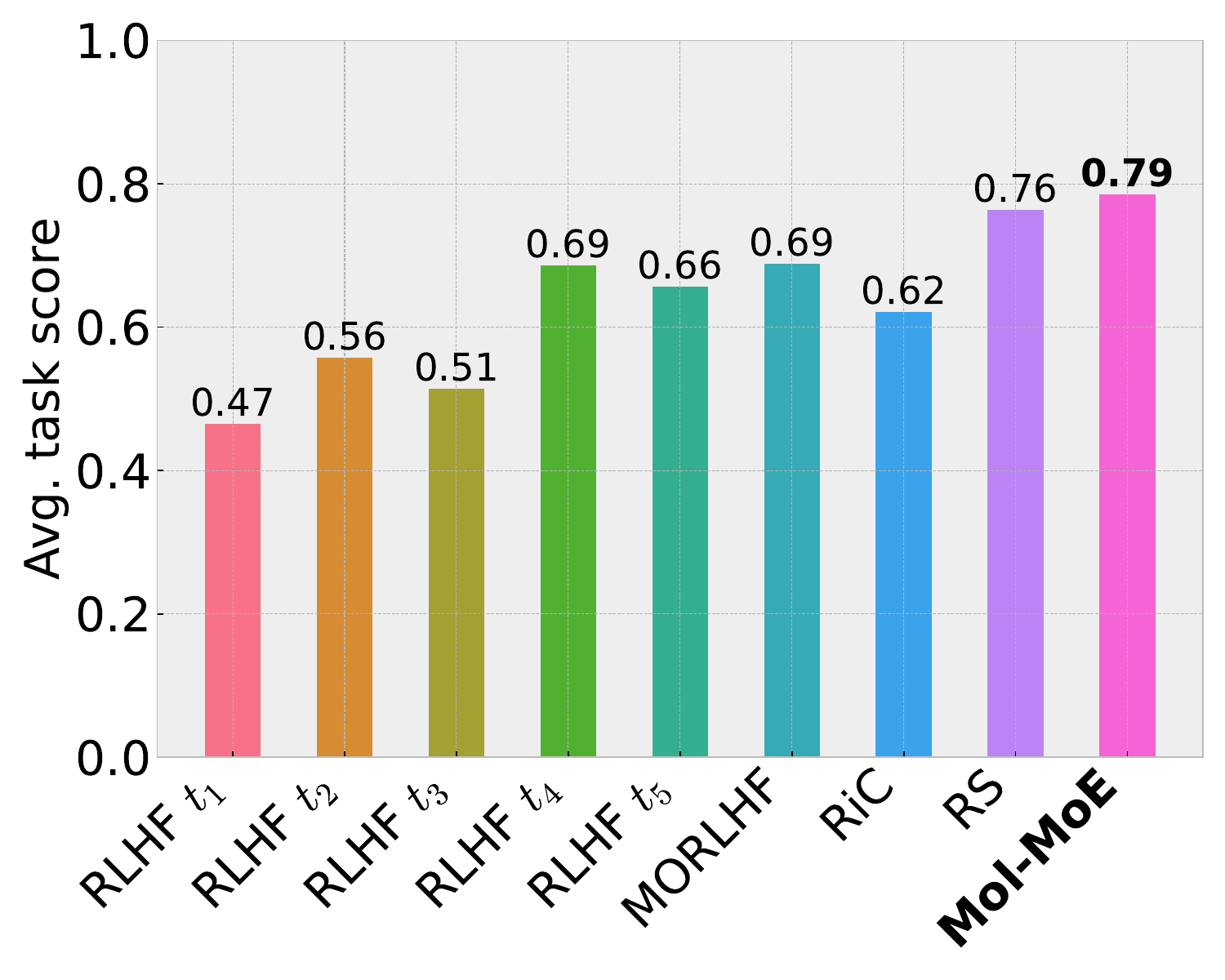}
    \vspace{-10pt}
    \caption{Out of distribution average property score by fine-tuning method. \molmoe outperforms multi-task models (MORLHF, RiC, RS) and by a significant margin also task experts (RLHF $t_1=$\texttt{JNK3}, $t_2=$\texttt{DRD2}, $t_3=$\texttt{GSK3$\beta$}, $t_4=$\texttt{CYP2D6}, $t_5=$\texttt{CYP2C19}). Higher scores for experts trained on $t_4, t_5$ indicate positive transfer.}
    \label{fig:bar_plot_maximization}
\end{figure}

Recent advances in large language models have enabled powerful generative capabilities for molecule design through text-based representations. We reformulate this sequence generation task as a multi-objective reinforcement learning (MORL) problem, following the blueprint of Reinforcement Learning from Human Feedback (RLHF, \citet{ouyang2022traininglanguagemodelsfollow}) but extending it to handle multiple competing objectives simultaneously. While there exists a rich literature in MORL \citep{Hayes2022}, traditional scalarization techniques require complete retraining to explore different trade-offs, making them computationally impractical for rapid iteration in drug discovery workflows.

To address this limitation, we follow the design choices adopted for rewarded soups \citep{ramé2023rewardedsoupsparetooptimalalignment} and more generally, model merging. Precisely, rewarded soups are obtained by linearly interpolating different fine-tuned versions of the same model, a simple form of task arithmetics \cite{yadav2023tiesmergingresolvinginterferencemerging}.
While rewarded soups show advantages over direct prompting \cite{wang2024conditionallanguagepolicygeneral}, their linear interpolation of rewards proves restrictive for exploring the complex trade-off landscape of drug design. We therefore introduce Molecule-MoE (\molmoe), a mixture-of-experts approach that enables dynamic steering of molecular generation across multiple objectives while maintaining computational efficiency through activation-space manipulation rather than full model retraining.

\section{Background}

\textbf{Deep learning and virtual screening}~ 
Virtual screening~(VS) is the process of selecting potential drug candidates entirely through computational methods -- in silico -- by leveraging large collections of known compounds to prioritize the most promising candidates for further study.
As public collections of chemical compounds expanded to billions of samples \cite{zinc22}, advances in Natural Language Processing (NLP) \cite{brown2020languagemodelsfewshotlearners, kaplan2020scalinglawsneurallanguage, ouyang2022traininglanguagemodelsfollow, pfeiffer2024modulardeeplearning} proved highly impactful. Sequence models, such as Transformers \cite{vaswani2023attentionneed}, can now learn from these vast datasets, enabling a range of downstream applications—including embeddings for virtual screening \cite{shen2022svsbisequencebasedvirtualscreening}, docking prediction \cite{zholus2024bindgptscalableframework3d}, and property optimization \cite{cavanagh2024smileyLlamamodifyinglargelanguage}.

\textbf{Property optimization with Reinforcement Learning}~ While a generative model trained on molecules reproduces the underlying distribution of compounds in the dataset, our goal extends beyond this: we aim to sample more compounds from promising regions of the chemical space by targeting specific areas of the learned molecular space that align with desired properties. Although the features describing molecules in this target region can be learned from wet lab discoveries, extracting high-quality labels is costly \cite{huang2021therapeuticsdatacommonsmachine}. This is why machine learning classifiers trained on this data are often used to provide approximate targets for downstream tasks \cite{ghugare2023searchinghighvaluemoleculesusing, tripp2022an} rather than relying on human annotators. This enables the alignment of language models in the molecular space with desired molecule properties through RLHF techniques, such as Direct Preference Optimization (DPO) \cite{cavanagh2024smileyLlamamodifyinglargelanguage}, Proximal Policy Optimization (PPO) \cite{schulman2017proximalpolicyoptimizationalgorithms}, or REINFORCE Leave-One-Out (RLOO) \cite{ahmadian2024basicsrevisitingreinforcestyle}.

The need to balance multiple criteria in real-world applications has motivated two primary approaches to multi-objective alignment in large language models. The first approach uses supervised fine-tuning (SFT) \cite{yang2024rewardsincontextmultiobjectivealignmentfoundation}, which has been applied to drug discovery tasks. The second approach employs model merging techniques and multi-objective RLHF (MORLHF) \cite{ramé2023rewardedsoupsparetooptimalalignment}, though its application to drug discovery remained unexplored prior to our work. Our experiments reveal fundamental limitations in both approaches: linear weight interpolation (model soups) fails to capture complex nonlinear relationships between objectives, while pareto-conditioning SFT methods \cite{yang2024rewardsincontextmultiobjectivealignmentfoundation} struggle with extrapolation beyond their training datasets.

Generative Flow Networks (GFlowNets) \cite{bengio2023gflownetfoundations} have emerged as a promising framework for molecular generation. Furthermore, theoretical work by \cite{tiapkin2024generativeflownetworksentropyregularized} has established their connection to entropy-regularized reinforcement learning by reframing GFlowNet, enabling connections to the rich multi-objective reinforcement learning literature. \cite{jain2023multi} has further demonstrated GFlowNets' effectiveness in balancing multiple criteria using preference-conditioning and scalarization techniques. Despite these successes, GFlowNets face inherent scaling challenges compared to large language models (LLMs), which benefit from extensive pre-training on massive text corpora. GFlowNets have yet to demonstrate the capacity to process or learn from token volumes on a similar scale.

\textbf{Desiderata}~ In the context of drug design, we aim to assess these approaches against two main criteria: quality of the generated samples and precision in following target preferences (steerability). Our analysis reveals that while model merging approaches show promise in combining multiple objectives, they suffer from two key limitations: (1) the inability to capture complex non-linear relationships between different molecular properties and (2) sub-optimal performance when precise control over individual properties is required. These limitations stem from the inherent constraints of linear interpolation in model merging, where the influence of each expert model is determined by fixed weights rather than as a by-product of a dynamic and input-dependent \textit{routing} decision. 

To address these challenges, we draw inspiration from Mixture of Experts (MoE) \cite{jiang2024mixtralexperts} architectures, which have demonstrated success in handling complex, multi-modal tasks by dynamically routing inputs to specialized expert networks \cite{komatsuzaki2023sparseupcyclingtrainingmixtureofexperts}. We further introduce \molmoe, a novel architecture built upon model merging that leverages the strengths of both approaches: we combine multiple expert MLP layers in a mixture of experts blocks \cite{zhou2022mixtureofexpertsexpertchoicerouting} and devise a routing training task to improve steerability. %
This architecture introduces dynamic, input-dependent weighting of various molecular property objectives, enhancing steerability while preserving the test-time advantages of model merging, rather than relying solely on training-time optimizations.

To summarize, we list our contributions:
\begin{itemize}[itemsep=2pt, parsep=0pt, topsep=2pt]
    \item We extend recent advancements in multi-objective RL—such as supervised fine-tuning (SFT) and model merging—to the domain of molecule design for drug discovery. We evaluate their performance in terms of both quality and controllability, as detailed in Sections \ref{maximization_analysis} and \ref{steerability_analysis}.
    \item We propose a novel approach based on model merging and mixture of experts: \molmoe, combining property-specific molecule LMs with a tuning task aimed at improving steerability (Section \ref{proposed_method}).
    \item We conduct a series of analyses to investigate several key factors: out-of-distribution performance, the relationship between merging hyperparameters and performance, the impact of the RLHF algorithm and model size, and the choice of merging technique (Section \ref{experiments} and Appendix \ref{supplementary_analyses}).
\end{itemize}
\section{Problem setup}
Drug design requires optimizing multiple molecular properties across diverse categories: ADME (absorption, distribution, metabolism, excretion) properties \cite{huang2021therapeuticsdatacommonsmachine}, drug-likeness scores, and binding affinities. For high-quality drug candidates, the number of relevant properties typically reaches around 10 \cite{fangwang2023cadme}, varying based on the stage of drug development, desired property granularity, and available domain knowledge. Additionally, as clinical trials progress, new data from diverse patient cohorts necessitates efficient updates to multiple property estimators \cite{fangwang2023cadme}. However, for the purpose of our experiments, we consider a stationary multi-objective MDP where these properties remain fixed.

\begin{figure*}[t]
    \centering
    \includegraphics[width=450pt]{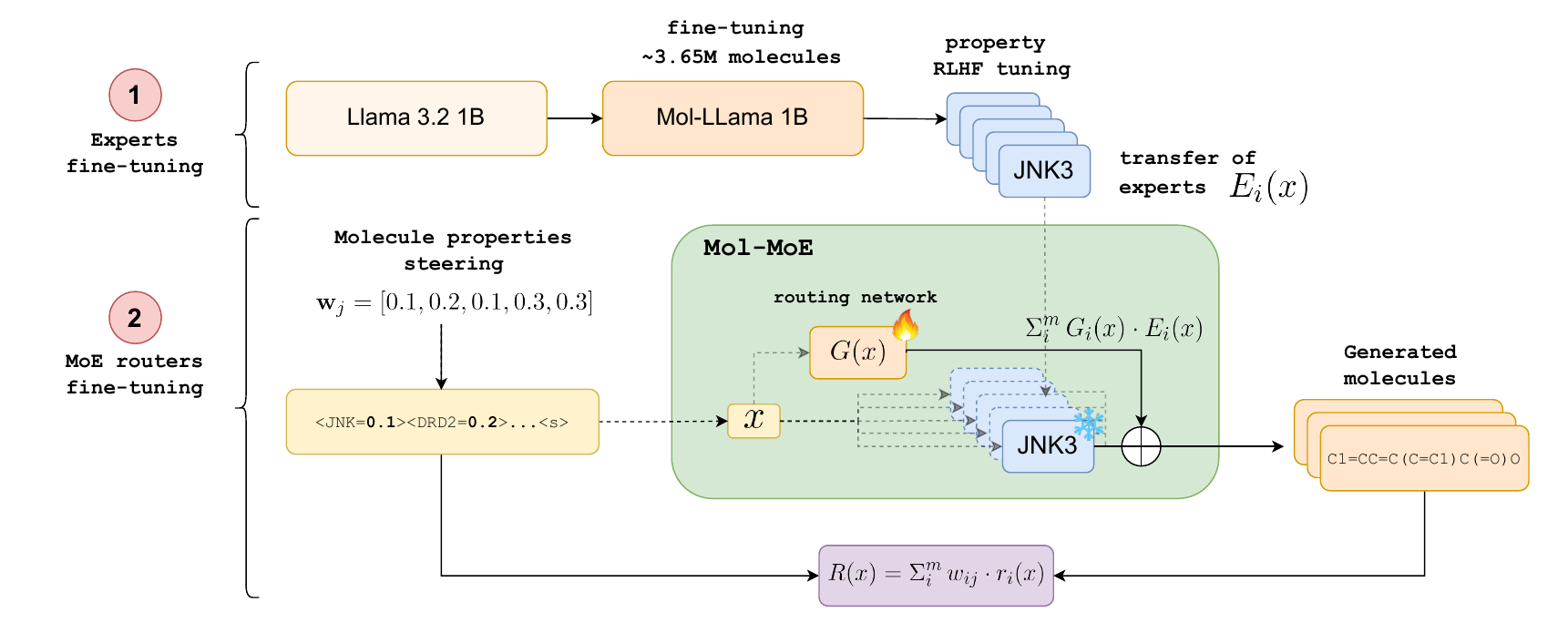}
    \caption{Pipeline illustration of \molmoe: a reference model is pre-trained on a large set of molecules; expert models are derived with RLHF-tuning on each desired molecule property; in the fine-tuned model, MoE blocks are added to combine the tuned expert layers, only the router networks are further tuned with the routing task.}
    \label{fig:pipeline}
\end{figure*}

\subsection{Multi-Objective RL}
We formulate drug design as a multi-objective Markov decision problem (MOMDP), where we aim to optimize a set of competing molecular properties represented 
by the reward vector $\mathbf{r}(x) = [r_1(x), ..., r_m(x)] \in \mathbb{R}^m$. Each objective is weighted by a preference vector $\mathbf{w}_j \in \mathcal{W} \subseteq \mathbb{R}^m$, where $\mathcal{W}$ is the space of valid preference vectors, and can be either learned (using preference modelling methods) or pre-specified. Let $f_{\boldsymbol{\theta}}: \mathcal{X} \rightarrow \mathcal{A}$ be a policy function mapping states to actions, parameterized by $\boldsymbol{\theta}$. Multi-objective RL (MORL) approaches can be categorized into two main design paradigms: \textit{single-policy} or \textit{multi-policy} methods. {Single-policy} methods train a unique policy parameterized by $\boldsymbol{\theta}^*$ that either generalizes across all possible preferences $\mathcal{W}$ or maximizes a specific preference vector $\mathbf{w}_j$. However, these approaches often yield suboptimal performance or fail to scale to complex tasks as the burden falls primarily on the network's generalization capacity and quality of the training distribution \cite{yang2019generalizedalgorithmmultiobjectivereinforcement}.

Conversely, \textit{multi-policy} methods instead maintain a set of policies, each parameterized by $\boldsymbol{\theta}^*_j \in \Theta$, and optimized for a specific preference vector $\mathbf{w}_j = [w_{j,1}, ..., w_{j,m}]$. In the most straightforward implementation, objectives are combined through scalarization:
\begin{align*}
    R_j(x) &= \sum_{i=1}^m w_{j,i} r_i(x) \,,\, \mathbf{w}_j \in \mathcal{W} \\
    \boldsymbol{\theta}^*_j &= \text{argmax}_{\boldsymbol{\theta} \in \Theta} \,  R_j(f_{\boldsymbol{\theta}}(x))
\end{align*}
The scalarized reward function $R_j: \mathcal{X} \rightarrow \mathbb{R}$ then reduces the multi-objective problem to a standard single-objective RL optimization for the given preference vector $\textbf{w}_j$. In our context, we refer to multi-policy MORL methods as \textit{Multi Objective RLHF} (MORLHF) since traditional approaches in this field tend to elicit the reward vector $\mathbf{r}$ from human feedback. Despite not requiring human annotation in our work, we maintain this terminology for consistency. In the molecular sequence modeling context, a policy parameterized by $\boldsymbol{\theta}$ and fine-tuned on a large molecular space can be optimized for a preference vector $\mathbf{w}_j \in \mathcal{W}$ using any alignment algorithm (e.g., RLOO, DPO, PPO), yielding $\boldsymbol{\theta}^*_j$. 

The multi-policy (MORLHF) setup faces two practical challenges: first, each preference vector $\mathbf{w}_j \in \mathcal{W}$ requires training a separate policy, making preference space exploration costly; second, the varying scales of rewards $r_i(x)$ necessitate careful scaling of preferences to ensure stable training, often requiring techniques like reward centering \cite{naik2024rewardcentering}. We analyze these challenges in Section \ref{scaling_tasks_analysis}.

\subsection{Expert Policy Interpolation}
Alternatively, \citet{ramé2023rewardedsoupsparetooptimalalignment} leverages Linear Mode Connectivity (LMC) \cite{frankle2020linearmodeconnectivitylottery} in Transformers to derive multi-task policies through linear interpolation of task-specialized policies. For a set of $m$ tasks, we first obtain specialized policies parameterized by $\boldsymbol{\theta}_i$ through single-task optimization. For each \textit{task} (property), we compute: 
$$\boldsymbol{\theta}_i = \text{argmax}_{\boldsymbol{\theta} \in \Theta} \,r_i(f_{\boldsymbol{\theta}}(x)) \quad \text{for } i \in \{1,\ldots,m\}$$
These task-specific policies form a \textit{model soup} (referred to as a \textit{rewarded soup}, RS, in this context). The key property of this collection is that the policy parameters $\{\boldsymbol{\theta}_i\}_{i=1}^m$ can be linearly combined using preference weights to yield a merged policy whose parameters $\boldsymbol{\theta}^*_j$ are obtained as:
$$\boldsymbol{\theta}^*_j = \sum_{i=1}^m w_{j,i} \boldsymbol{\theta}_i \quad \text{where } \mathbf{w}_j \in \mathcal{W}$$
This merged policy aims to reflect the relative importance of each property as specified by the preference vector $\mathbf{w}_j \in \mathcal{W}$.
This method features good steerability \cite{wang2024conditionallanguagepolicygeneral} while benefiting from the rich advances in model merging
\cite{yadav2023tiesmergingresolvinginterferencemerging, davari2024modelbreadcrumbsscalingmultitask} meant to mitigate task interference. Remarkably, generating policies for new preference vectors requires only parameter interpolation, without additional training, as the multi-task policies are computed offline.

This feature is particularly relevant in the context of molecule design, where numerous properties must be incorporated over time. The relative importance of these properties is often difficult to predict in advance, especially as clinical data becomes available and may introduce distributional shifts. This necessitates using flexible methods that can efficiently adapt to evolving preferences without requiring costly retraining. 

One of the main drawbacks of this approach, which we aim to address in this work, is that the relationship between the defined preference set \(\textbf{w}_j \in W\) and the corresponding task performance is not always linear. As a result, predicting how a preference vector should be translated into an interpolation weight for the rewarded soup mode is challenging. To address this, we experiment with learning a mapping function, such as \(w_{ij}' = z(w_{ij})\) \citep{yang2024rewardsincontextmultiobjectivealignmentfoundation}, to ensure that models derived from rewarded soups are steerable—i.e., capable of responding to the given preference vector as intended. However, we demonstrate that a more effective approach is to learn an input-dependent weighting function through a dedicated \textit{routing} task, as implemented in our proposed \molmoe architecture.

\subsection{Supervised Learning}

\citet{yang2024rewardsincontextmultiobjectivealignmentfoundation} propose a fully supervised approach inspired by imitation learning \cite{zare2023surveyimitationlearningalgorithms} and aligned with Pareto-conditioning techniques \cite{reymond2022paretoconditionednetworks, chen2021decisiontransformerreinforcementlearning}. This framework provides the desired property scores as a prefix when conditioning the model on a target molecule. The underlying assumption is that causal language modeling enables the Transformer to learn the relationship between input rewards (interpreted as a state) and the target molecule sequence (interpreted as an action). This mirrors the way decision transformers \cite{chen2021decisiontransformerreinforcementlearning} learn to associate trajectory returns with optimal actions. The model is initially tuned on a large set of known molecules; then a filtering process defines a subset of Pareto-optimal samples augmented by the model generations, and finally, the network undergoes a second fine-tuning step on the high-quality dataset. This approach is identified as \textbf{Rewards in Context} (RiC, \citet{yang2024rewardsincontextmultiobjectivealignmentfoundation}), it achieves comparable or superior sample quality wrt. MORLHF and RS. However, as this method heavily relies on the quality of the training dataset, it can be prone to overfitting. With an ablation analysis in Section \ref{ood_analysis}, we show that models trained with RiC fail to generate molecules better than the training examples, and the fraction of repeated generations is high. In the context of drug design, our goal is not only to replicate the statistical patterns of the training data but also to surpass the quality of the imitation data—generating novel, high-quality compounds that were not present in the training set.
While RiC offers test-time steerability similar to RS approaches, its inability to generalize (or "maximize") beyond the training data motivates our development of the \molmoe approach to combine the strengths of both methodologies while avoiding their pitfalls.

\section{\molmoe: Molecule-Mixture of Experts}
\label{proposed_method}

Linear interpolation methods in model soups and reward scalarization share surface-level similarities, but their underlying mechanisms differ in important ways. The relationship between preference vectors \(\mathbf{w}_{ij}\) for policy parameters \(\boldsymbol{\theta}_i\) and model soup coefficients is neither equivalent in magnitude nor necessarily linear.

Our correlation analysis (Appendix \ref{section:correlation_analysis}) reveals that simply assigning equal weights to policies that maximize individual tasks does not guarantee optimal multi-task performance. This limitation has led researchers to explore more sophisticated approaches, such as iterative and gradient-based optimization methods, to find optimal parameter combinations \cite{huang2024lorahubefficientcrosstaskgeneralization, prabhakar2024lorasoupsmergingloras}.

Furthermore, the challenge becomes more complex with increasing network depth, as the semantic meaning of representations evolves across layers. Recent work has proposed systematic approaches to address this issue. Model Breadcrumbs \cite{davari2024modelbreadcrumbsscalingmultitask} computes update masks based on the distribution of task-specific parameters to mitigate interference patterns. TIES \cite{yadav2023tiesmergingresolvinginterferencemerging} extends this by introducing element-wise merging rules based on the relative magnitudes and signs of corresponding parameters - when parameters from different tasks have opposing signs above a threshold, it preserves the dominant direction rather than averaging them.

While these approaches introduce data-driven methods to improve parameter merging through similarity analysis and interference detection, they ultimately rely on predetermined merging rules. To move beyond fixed combination strategies, we propose leveraging the \textbf{Mixture of Experts} \cite{jiang2024mixtralexperts} architecture, which learns to route inputs to specialized components dynamically. 

In each Transformer layer $\ell$, the feed-forward layers after attention are replaced with a MoE-block consisting of $m$ expert networks $E_i: \mathbb{R}^d \rightarrow \mathbb{R}^d$, where each $E_i(\mathbf{x}), \, i \in \{1, \hdots, m\}$ is implemented as a multi-layer perceptron (MLP). The routing mechanism is controlled by a router network $f_{\boldsymbol{\theta}_g}: \mathbb{R}^d \rightarrow \mathbb{R}^m$ with parameters $\boldsymbol{\theta}_g$, which computes routing weights for each expert.
For each input token $\mathbf{x} \in \mathbb{R}^d$, the router outputs a probability distribution $\mathbf{G}(\mathbf{x}) \in \mathbb{R}^m$ over experts:
\begin{equation*}
\mathbf{G}(\mathbf{x}) := \operatorname{softmax}(\operatorname{topk}(f_{\boldsymbol{\theta}_g}(\mathbf{x})))\enspace,
\end{equation*}
where $\operatorname{topk}$ selects the $K$ largest logits and sets the rest to $-\infty$, ensuring that each token is processed by at most $K$ experts.
Here, $K$ represents the number of experts assigned to each token. In our setup, we set $K = m$, as the number of specialists is known a priori and corresponds directly to the properties we aim to optimize. This ensures that all specialist layers contribute to constructing the target molecule, each addressing a specific property. Additionally, we include the initialization model—trained on a large dataset of molecules—as an extra expert to provide regularization \cite{royer2023revisitingsinglegatedmixturesexperts}. 
The final output for each token combines the expert outputs weighted by their corresponding routing probabilities:
\begin{equation*}
    \hat{\mathbf{x}} = \sum_{i=1}^{m} \, G_i(\mathbf{x}) \cdot E_i(\mathbf{x})\enspace,
\end{equation*}
where $G_i(\mathbf{x})$ denotes the $i$-th component of the routing vector $\mathbf{G}(\mathbf{x})$.

\subsection{Preference-Guided Router Training}

Our method employs a data-driven strategy for dynamic model merging via learned routing. Given a set of expert policies \(\{E_i\}_{i=1}^m\) and a preference vector \(\mathbf{w}_j \in \Delta^{m-1}\), where \(\Delta^{m-1}\) represents the probability simplex (i.e., \(\sum_{i=1}^m w_{j,i} = 1\) and \(w_{j,i} \geq 0\) for all \(i\)), we encode preferences into the input prompt using a structured format such as ``\texttt{<JNK3=w1><DRD2=w2>...<s>}". This prompted input is then processed through our routing architecture to learn an input-dependent routing function \(\mathbf{G}(\mathbf{x}; \boldsymbol{\theta}_g)\), which determines how to combine the outputs of the experts. Similar to the preference weights, the routing weights also form a probability distribution over the experts, i.e., \(\sum_{i=1}^m G_i(\mathbf{x}) = 1\) and \(G_i(\mathbf{x}) \geq 0\) for all \(i\).

The router parameters \(\boldsymbol{\theta}_g\) are trained using reinforcement learning across a distribution of preference vectors. For each training episode, we sample \(\mathbf{w}_j\) from a set of possible preferences \(\{\mathbf{w}_j\}_{j=1}^N \subset \Delta^{m-1}\), encode it into the prompt, and use this to define the episode's scalarized reward.
By training over the distribution of preference vectors, we optimize the expected reward across all possible preference configurations:
\[
\mathbb{E}_{\mathbf{w}_j \sim P(\Delta^{m-1})} \left[ R_j(\mathbf{x}) \right] = \mathbb{E}_{\mathbf{w}_j \sim P(\Delta^{m-1})} \left[ \sum_{i=1}^m w_{j,i} \cdot r_i(\mathbf{x}) \right].
\]
This formulation ensures that the router learns to interpret preferences from the prompt and dynamically adjusts the contributions of each expert to align with these preferences, without requiring post-hoc calibration. In contrast, static merging approaches like Model Breadcrumbs or TIES must solve an additional optimization problem to match their fixed merging rules with desired preference weights, often resulting in suboptimal trade-offs due to their predetermined combination strategies.

Our approach shares the high-level goal of preference-conditioned generation with Conditional Language Policy (CLP) \cite{wang2024conditionallanguagepolicygeneral}, but differs significantly in implementation. While CLP fine-tunes the entire model to condition its policy on user preferences directly, our method leaves the underlying expert policies unchanged. Instead, we introduce a routing mechanism that dynamically combines the outputs of pre-trained experts based on preferences encoded in the input prompt. By embedding preferences directly into the prompt using a structured format, our architecture provides a flexible and robust way to condition model behavior using natural language while minimizing sensitivity to prompt variations.

\begin{figure*}[t]
    \centering
    \begin{minipage}{0.7\textwidth}
        \centering
        \includegraphics[width=0.9\linewidth]{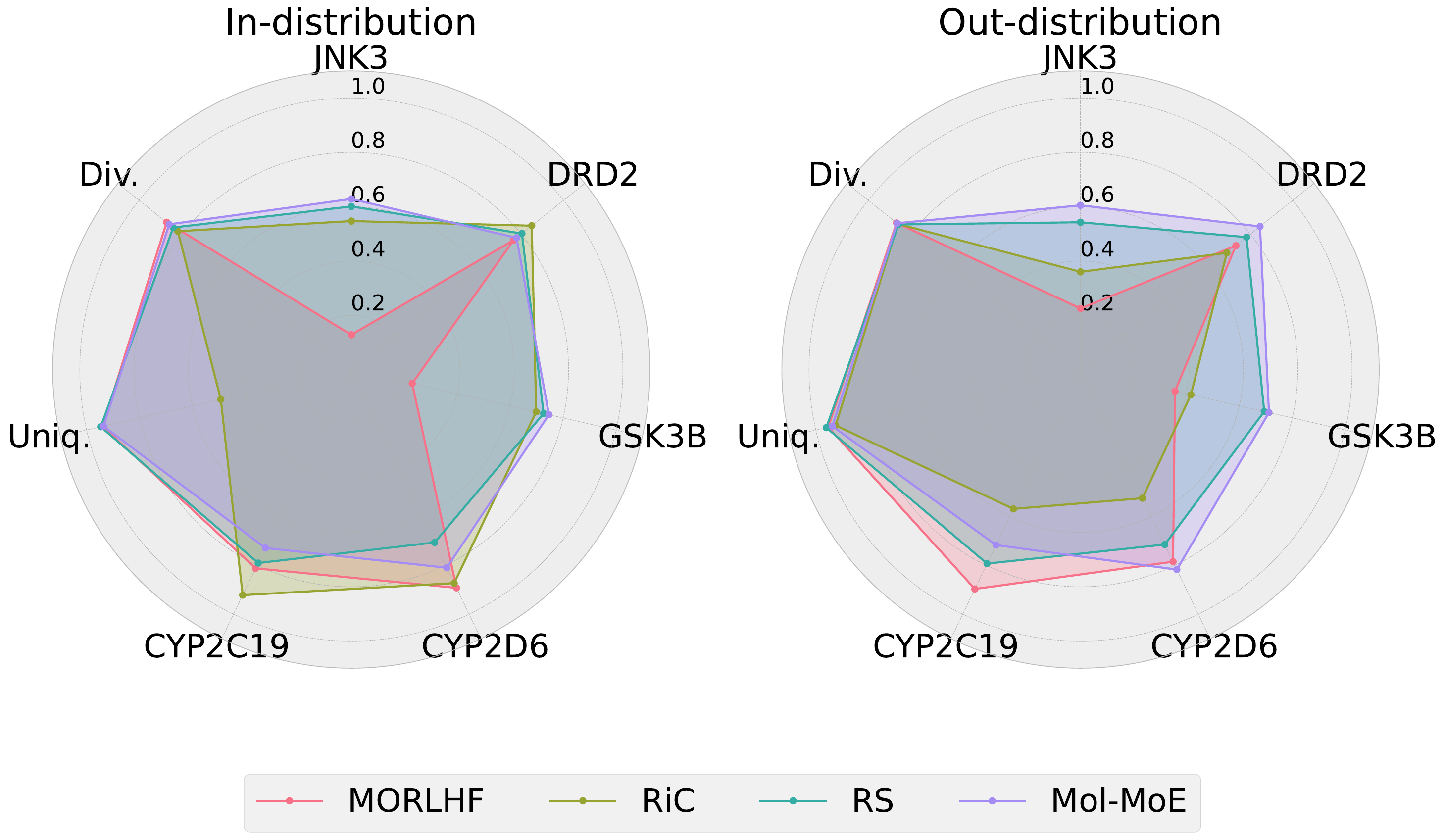}
    \end{minipage}%
    \caption{Average best task scores achieved by models trained on the full dataset (left) or with held-out high-quality samples (right), by varying tuning method. \molmoe outperforms the baselines particularly out of distribution, where RiC fails to improve beyond on the training examples. Additionally, MORLHF fails to learn more than three tasks.}
    \label{fig:maximization_evaluation}
\end{figure*}

Unlike traditional Mixture of Experts (MoE) architectures \cite{jiang2024mixtralexperts}, which train experts jointly from scratch, our method leverages pre-trained task-specific policies through activation-level routing—a process referred to as ``upcycling'' \cite{komatsuzaki2023sparseupcyclingtrainingmixtureofexperts}. This approach eliminates the need for explicit expert specialization, as each expert is already optimized for a specific molecular property, while the router learns to blend their outputs based on input preferences adaptively.

We choose RLOO \cite{ahmadian2024basicsrevisitingreinforcestyle} as training algorithm, after an empirical evaluation across alternatives (Section \ref{rlhf_analysis}). The objective functions $r_i(x)$ consist of MLP property classifiers trained on clinical data \citet{huang2021therapeuticsdatacommonsmachine}: the models estimate the likelihood of an input molecule to satisfy the target property, similarly to examples observed in the past; the assumption is that the training set is sufficiently large for these models to well extrapolate for similar structures.

\section{Experiments}
\label{experiments}

In this section, we evaluate \molmoe against scalarized multi-objective RLHF (MORLHF) variants under a fixed scalarization with uniform weights, Rewarded Soups (RS), and Rewards in Context (RiC). Our evaluation is based on two main criteria: \textit{maximization}, that is sample quality over the considered tasks (Section~\ref{maximization_analysis}); \textit{steerability}, how effectively each method tunes the model to follow conditioning preferences $\mathbf{w}_j \in \mathcal{W}$ at test-time (Section~\ref{steerability_analysis}). 

We report further details concerning the implementation details in Appendix \ref{implementation_details} and supplementary analyses supporting our design choices in Appendix \ref{supplementary_analyses}. In Appendix \ref{section:correlation_analysis} we further investigate the relationship between conditioning preferences $\mathbf{w}_j$ and property scores $r_i(x)$.

\subsection{Experimental Setting}
\textbf{Molecule pre-training}~~We assemble a training set of approximately 3.65 million molecule sequences by merging datasets from CHEMBL \cite{Gaulton2012}, ZINC 250k \cite{doi:10.1021/acs.jcim.5b00559}, and MOSES \cite{10.3389/fphar.2020.565644}. Using the Llama 3.2 tokenizer \cite{grattafiori2024Llama3herdmodels}, this dataset translates to approximately 151 million training tokens, with an average of 41.44 tokens per molecule. We represent molecule sequences in the SMILES format \cite{doi:10.1021/ci00057a005}, allowing us to work in text space and ensuring compatibility with existing foundational models. 

To create the base molecule language model, \textsc{Mol-Llama} we fine-tune the Llama 3.2 1B model on the 3.65M dataset for causal language modeling, where the model learns to predict each token based on all previous tokens in the sequence. The scaling analysis detailed in Section~\ref{scaling_analysis} informs the model size choice.

\begin{figure*}[t]
    \centering
    \includegraphics[width=1.0\linewidth]{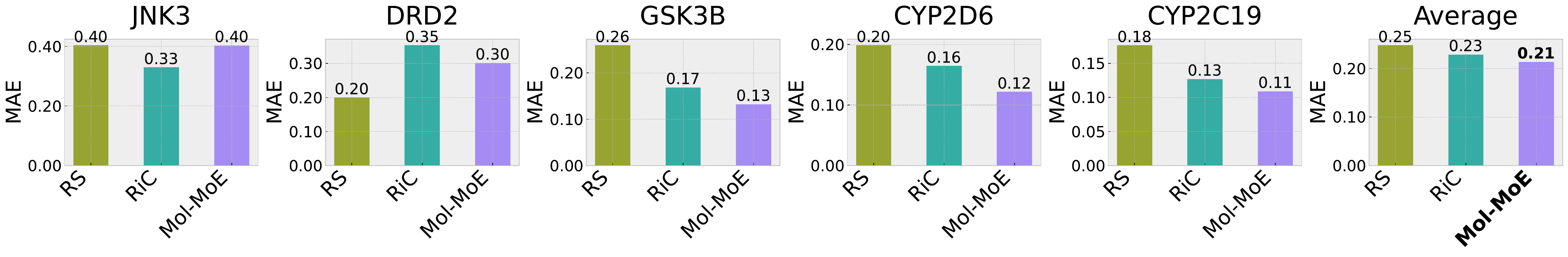}
    \vspace{-20pt} 
    \caption{Steerability error measured as Mean Absolute Error (MAE) between the conditioning the measured molecule properties. \molmoe is overall more precise than RiC and RS, particularly on \texttt{GSK3$\beta$, CYP2D6, CYP2C19}.}
    \label{fig:steerability_evaluation}
\end{figure*}

\textbf{Property RLHF fine-tuning}~~
We derive {\molmoe} from the pipeline shown in Figure~\ref{fig:pipeline}, implementing our approach using MergeKit \cite{goddard2025arceesmergekittoolkitmerging}. First, we obtain five experts by fine-tuning five copies (see appendix \ref{details_experts_rlhf}) of the base model on five molecule properties separately, using the scoring functions provided by \citet{huang2021therapeuticsdatacommonsmachine}, i.e. receptor binding scores.

To combine these experts effectively, we create MoE blocks where property-specific patterns guide the routing. Specifically, we initialize the router weights using activation patterns obtained by passing each property name (e.g., \texttt{JNK3}, \texttt{DRD2}) through the model. This initialization ensures the routers begin with a strong bias toward directing inputs to their corresponding expert models. After initialization, we train only the router networks using RLOO~\cite{ahmadian2024basicsrevisitingreinforcestyle}, keeping all other layers frozen. Training details are reported in Appendix \ref{details_moe_rlhf}.

\subsection{Maximizing molecule properties} 
\label{maximization_analysis}

We evaluate different model tuning approaches across seven metrics. Five are target molecule properties from the Therapeutics Data Commons (TDC) \cite{huang2021therapeuticsdatacommonsmachine}: \texttt{JNK3}, \texttt{DRD2}, \texttt{GSK3$\beta$}, \texttt{CYP2D6} and \texttt{CYP2C19}. We additionally evaluate on \textit{diversity} -- which we define as the average pairwise Tanimoto distance between generated molecules \cite{ghugare2023searchinghighvaluemoleculesusing} -- and \textit{uniqueness}, which is the fraction of unrepeated samples.

To evaluate each method's peak performance capability, we generate $2,048$ molecules across $20$ different random seeds with each tuned model. We identify the molecule with the highest average score across all properties for each set,  then average these peak scores across all seeds to obtain our final performance metric.
For fair comparison across methods, we configure MORLHF and RS with uniform weights ($w_{ij} = \frac{1}{m} = 0.2$ for each of the $5$ properties). RiC uses maximum target values ($\hat{r}_{i} = 1.0$) in its prompt format: \texttt{<JNK3=1.00><DRD2=1.00>...<s>}. {\molmoe} uses the same prompt format but with normalized preference weights ($\hat{w}_{ij} = \frac{w_{ij}}{\sum_{k=1}^{m} \, w_{kj}}$).

Figure \ref{fig:bar_plot_maximization} shows that single-task RLHF achieves suboptimal performance across objectives, though it demonstrates positive transfer effects, particularly from \texttt{CYP2D6} and \texttt{CYP2C19} to \texttt{DRD2} and \texttt{GSK3$\beta$}. MORLHF performs comparably to single-task tuning, but notably optimizes only a subset of the considered tasks rather than the full objective set. The base model fine-tuned with RiC excels in-distribution, performing comparably or better than RS and \molmoe for \texttt{DRD2} and \texttt{CYP2C19}. However, its performance degrades substantially when evaluating on out-of-distribution high-quality molecules. In contrast, the Rewarded Soups (RS) approach maintains robust performance even out of distribution, leveraging experts explicitly trained to maximize individual objectives. \molmoe emerges as the strongest performer, achieving superior results in both scenarios and significantly outperforming RiC and MORLHF, particularly in out-of-distribution cases.

We further observe that with temperature $t = 1.0$, all the models do not repeat sequences from the training set, leading to a maximum novelty score with the metric implemented by \cite{huang2021therapeuticsdatacommonsmachine}. Tabular results are reported in Table \ref{tab:max_points}.

\subsection{Measuring steerability}
\label{steerability_analysis}

To evaluate steerability, we generate a test set by sampling 20 preference vectors from a 5-dimensional Dirichlet distribution, yielding $W \in \mathbb{R}^{20 \times 5}$. For each preference vector, we sample 2,048 molecules. We then compute steerability error for each molecular property. For RiC, which directly predicts absolute reward values, the error is measured against the prompted rewards:
\[\operatorname{MAE}(x) = \frac{1}{m} \sum_{i=1}^{m} (r_i(x) - \hat{r}_i)\]
For RS and \molmoe, we instead measure how well the generated molecules' properties match the requested preference proportions. This is done by comparing the requested preference weights against the normalized property scores:
\[\operatorname{MAE}(x) = \frac{1}{m} \, \sum_i^{m} \, \left(w_{ij} - \frac{r_i(x)}{\sum_{k=1}^{m} \, r_k(x)} \right) \,,\, \textbf{w}_j \in \mathcal{W}\]
where $w_{ij}$ represents the requested preference weight for property $i$, and the fraction $\frac{r_i(x)}{\sum_l^{m} \, r_l(x)}$ represents the actual proportion of property $i$ in the generated molecule relative to all properties.

Figure~\ref{fig:maximization_evaluation} and Figure~\ref{fig:steerability_evaluation} report the performance of different methods in terms of maximization and steerability metrics, showing that \molmoe improves over baselines in both evaluation criteria.

We observe that \texttt{JNK3} is a hard task in both maximization (Figure \ref{fig:maximization_evaluation}) and steerability (Figure \ref{fig:steerability_evaluation}), as all three methods achieve comparable scores. In \texttt{GSK3$\beta$}, \texttt{CYP2D6}, \texttt{CYP2C19}, \molmoe achieves the lowest error across properties, suggesting that the routing networks successfully learned to aggregate the contribution from different property experts as instructed.

\subsection{Out of distribution behavior}
\label{ood_analysis}

To rigorously test whether our models can discover novel high-quality molecules rather than imitate known ones, we create a challenging evaluation setup: we remove all high-scoring molecules (\texttt{$r_i(x) \geq$ 0.6} for any property $i$) from the pretraining dataset, leaving only $\sim 2.12$M molecules ($58 \%$). In Figure \ref{fig:maximization_evaluation} and Table \ref{tab:ood_max_points}, we observe a significant decline in performance for RiC, based on SFT, for which the model does not generate molecules better than the training distribution. Conversely, policies trained with \molmoe and RS maintain performances comparable to the in-distribution setup, while with MORLHF, the network optimizes for only a subset of the objectives. We thus infer that RiC, based solely on increasing the likelihood of high-quality training samples, poorly improves out of distribution. Conversely, RS and \molmoe successfully discover molecules with scores above 0.6, demonstrating their ability to explore and optimize in regions of chemical space beyond their training distribution.

\subsection{Scaling by number of properties}
\label{scaling_tasks_analysis}

\begin{figure}[t]
    \centering
    \hspace*{-0.25in}
    \includegraphics[width=0.9\linewidth]{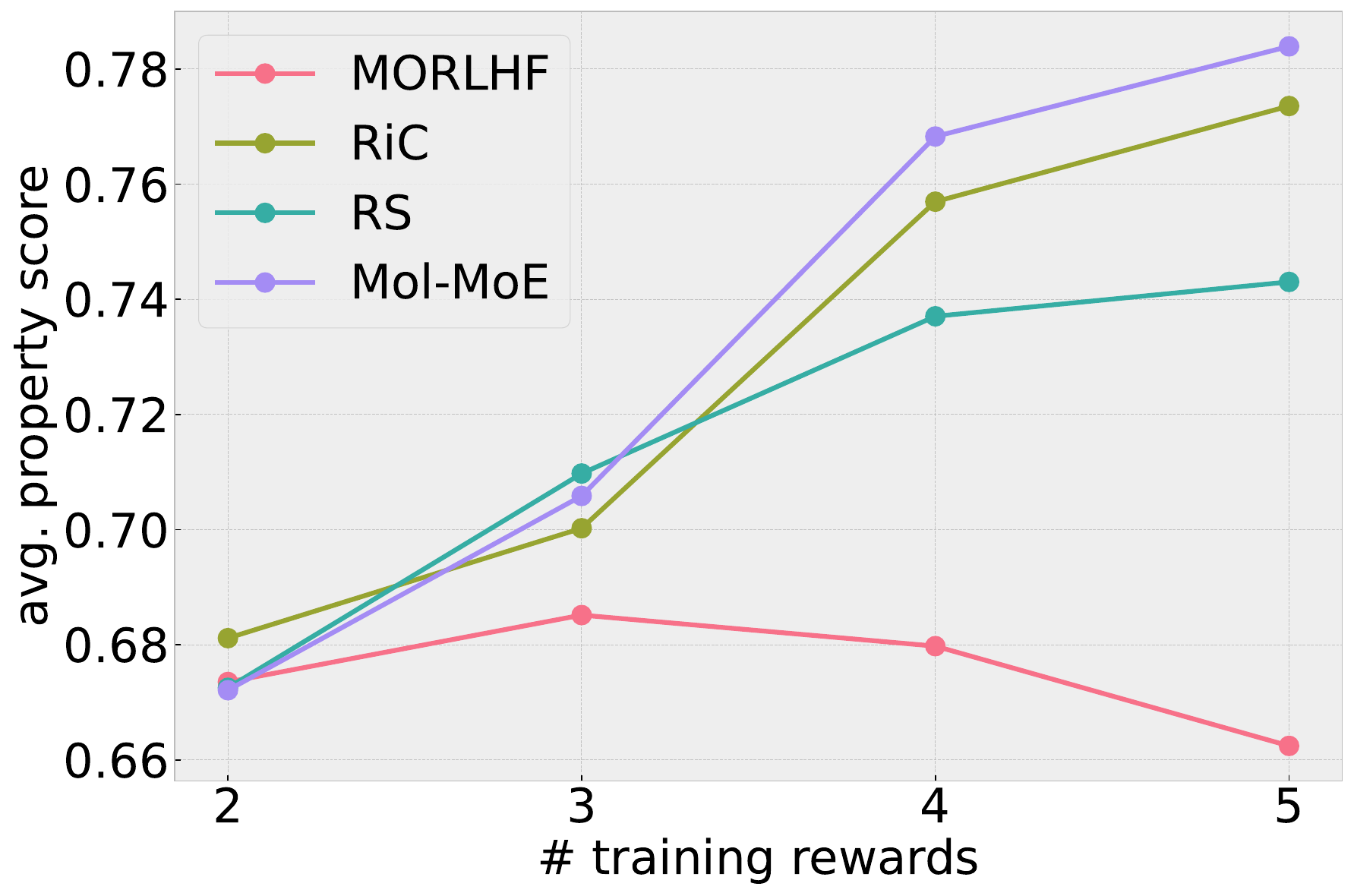}
    \caption{Average score on all mocule properties by number of training objectives by tuning method. \textbf{\molmoe} benefits the most from increasing reward signals, particularly wrt. RS. Conversely, MORLHF fails beyond three tasks, coherently to what observed in Figure \ref{fig:maximization_evaluation}.}
    \label{fig:tasks_scaling_analysis}
\end{figure}

We investigate how model performance scales with the number of target properties. Using identical configurations, we train each method (MORLHF, RS, RIC, \molmoe) on progressively larger subsets of properties: $m \in [2, 3, 4, 5]$. For each subset, we evaluate performance by averaging task scores across $2,048$ generated molecules. Additional properties should improve generation quality by providing more guidance to the models. Figure \ref{fig:tasks_scaling_analysis} shows that \molmoe, RiC and RS successfully leverage additional properties, with \molmoe achieving the highest quality. However, MORLHF's performance degrades when optimizing more than three properties ($m > 3$). This indicates that directly optimizing a scalarized reward becomes intractable as the number of properties increases, likely due to compounding complexity and interference between objectives. In contrast, \molmoe's architecture separates concerns: experts can stably optimize individual properties while the router manages interference between them.

\section{Conclusion}  
This work approaches molecule generation through language modeling, framing drug design as a multi-objective optimization problem. We evaluated three existing approaches: supervised learning through Rewards in Context (RiC), multi-objective reinforcement learning (MORLHF), and model merging via Rewarded Soups (RS). Each showed distinct limitations: RiC struggled with novel generation and out-of-distribution performance, MORLHF failed to optimize all objectives simultaneously, and RS lacked precise control over property trade-offs.

To address these limitations, we introduced \molmoe, a mixture-of-experts architecture that combines the strengths of expert specialization with preference-guided routing. A key contribution of our work is the novel preference-guided routing objective, which trains the router to dynamically adjust the contributions of expert models based on input preferences. This mechanism enables better calibration between preference weights and model outputs, resulting in more precise steerability and finer-grained control over molecular properties. While RiC and RS demonstrated strong in-distribution performance, \molmoe maintains this performance on novel compounds while offering superior control over property trade-offs.

However, our approach assumes that molecular properties can be optimized independently and combined through linear preference weighting, an assumption that breaks down in real-world drug design. Molecular property interactions are often non-monotonic: optimizing for one objective may degrade another, as seen in drug design where improving a compound's solubility can reduce its membrane permeability, or increasing potency may lead to unintended toxicity \cite{Bickerton2012, Leeson2007}. 
Future work should explore graph-based routing architectures or geometric deep learning approaches that explicitly model these interactions, allowing the router to learn and exploit the structured relationships between molecular properties rather than assuming simple additive effects.

Finally, to fully realize the potential of \molmoe in molecular design, future work should consider scaling the approach to billion-scale molecular datasets \cite{zinc22}. This will enable the model to uncover novel molecular trade-offs, generalize to a broader range of chemical spaces, and efficiently adapt to new optimization objectives -- all while avoiding the need for costly retraining.

\pagebreak
\bibliographystyle{icml2025}

\newpage
\appendix
\onecolumn

\section*{Appendix}

\section{Implementation details}
\label{implementation_details}

\subsection{Molecule fine-tuning}
\label{details_base_fine_tuning}
We fine-tune the base model ``Mol-Llama 1B'' for causal language modeling on $\sim$3.65M molecules, no conditioning information are provided: we add segmentation tokens to structure the model outputs, \texttt{<s> C1=CC=C(C=C1)C(=O)O </s>}. We consider a standard learning rate $\eta = 2e-5$ (with cosine scheduling), batch size $1024$ with gradient accumulation; the model is trained for 1 epoch.

\subsection{Experts fine-tuning}
\label{details_experts_rlhf}
We employ RLOO~\cite{ahmadian2024basicsrevisitingreinforcestyle} to fine-tune the different experts, rewarding each model on $100,000$ generations prompted with the start of sequence token \texttt{<s>}; batch size of $512$ samples; a constant learning rate of $\eta = 1e-7$; a KL coefficient of $\beta = 0.2$ to prevent policy collapse. We automatically assign a zero reward score to invalid SMILES string.

\subsection{Preference-Guided Router Training}
\label{details_moe_rlhf}

We employ RLOO~\cite{ahmadian2024basicsrevisitingreinforcestyle} to train the router networks in \molmoe while keeping the rest of the model layers frozen. The model is rewarded on $100,000$ generations prompted with a conditioning prompt encoding the preferences $w_{ij} \in \textbf{w}_j \,, \textbf{w}_j \in \mathcal{W}$ in the form ``\texttt{<JNK3=w1><DRD2=w2>...<s>}". We choose a fixed batch size of $128$ samples; learning rate $\eta = 1e-3$; KL coefficient $\beta = 5e-2$.

\subsection{Sampling parameters}
\label{details_sampling_params}

We keep the sampling parameters constant across the experiments: temperature $\text{t} = 1.0$; nucleus sampling as default strategy with $\text{top\_p} = 0.9$; max output length of $64$ tokens.

\pagebreak
\section{Supplementary analyses}
\label{supplementary_analyses}

\subsection{Choice of the RLHF algorithm}
\label{rlhf_analysis}

\begin{figure}[H]
    \centering
    \hspace*{-0.25in}
    \includegraphics[width=0.4\linewidth]{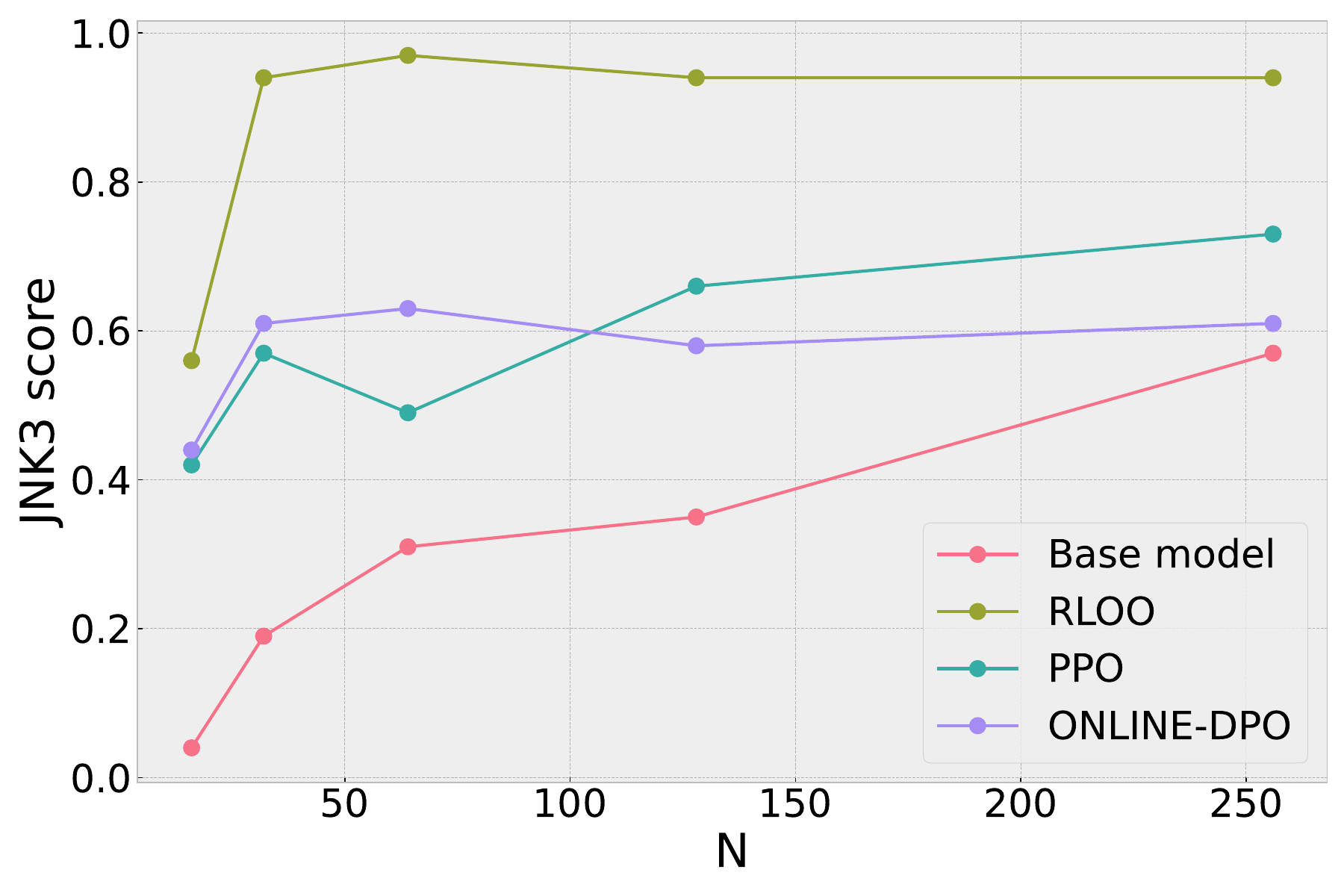}
    \caption{Task performance at sampling size by RL tuning method. Overall, RLOO surpasses alternative methods by a significant margin, while PPO and Online-DPO achieve comparable performance.}
    \label{fig:rlhf_algorithm_analysis}
\end{figure}

We compare three algorithms widely applied in RLHF: PPO \cite{schulman2017proximalpolicyoptimizationalgorithms}, RLOO \cite{ahmadian2024basicsrevisitingreinforcestyle} and Online-DPO \cite{guo2024directlanguagemodelalignment}. Each method is applied to fine-tune the base molecule sequence model on the non-trivial \texttt{JNK3} property. For each variant, we sample a growing set of $N$ molecules from the model and take the best-of-N property scores. We consider a moderate coefficient $\beta = 0.2$ for the KL divergence. 

In Figure \ref{fig:rlhf_algorithm_analysis} we observe the model generates qualitatively the best molecules with RLOO, which is in line with comparisons reported by \citet{ahmadian2024basicsrevisitingreinforcestyle}. Overall, we observe RL fine-tuning significantly improves the output distribution of molecules wrt. the base model, as it is possible to generate equivalently good compounds with $\sim 10$x less samples.

\subsection{Choice of the merging technique}
\label{merging_analysis}

\begin{table}[H]
    \centering
    \begin{tabular}{lccc}
        \toprule
        Method & Avg. score & $w_1$ & $w_2$ \\
        \midrule
        breadcrumbs & 0.91 & 0.4 & 0.6 \\
        dare\_linear & 0.91 & 0.4 & 0.6 \\
        ties & 0.90 & 0.5 & 0.5 \\
        della & 0.90 & 0.4 & 0.6 \\
        linear & 0.89 & 0.7 & 0.3 \\
        task\_arithmetic & 0.89 & 0.7 & 0.3 \\
        \bottomrule
    \end{tabular}
    \caption{Merging methods by average task score at optimal interpolation weights for \texttt{JNK3}, \texttt{DRD2}. Whiel Model breadcrumbs and DARE achieve the highest average property score, we do not observe a significant difference from the alternative methods.}
    \label{table:merging_techniques_compared}
\end{table}

A major motivation underlying this work is to explore the efficacy of weight interpolation in MORL. Combining linearly expert policies, as in Rewarded Soups, is the simplest form of merging, yet successful applications spawned a research line on ``task arithmetics'' \cite{yadav2023tiesmergingresolvinginterferencemerging, davari2024modelbreadcrumbsscalingmultitask, yu2024languagemodelssupermario, goddard2025arceesmergekittoolkitmerging}. We test different merging techniques in our experimental setup: from the base molecule model, we derived 2 experts fine-tuned with RLOO on \texttt{JNK3} and \texttt{DRD2} respectively; for each merging method, we performed a hyperparameter search for the optimal weight combination $[w_1, w_2] \in [0,1] \times [0,1]$ by average task score over $2,048$ generated samples. In Table \ref{table:merging_techniques_compared} we observe our policies merged with Model Breadcrumbs \cite{davari2024modelbreadcrumbsscalingmultitask} achieve the highest scores average; we notice that for this solution, a lower weight is assigned to \texttt{JNK3}, which is considered as a hard task. Overall, no significant improvement is observed by varying the merging method, which aligns with the experiments conducted by \citet{yadav2024mattersmodelmergingscale}.

\subsection{Choice of the model size}
\label{scaling_analysis}

\begin{figure}[H]
    \centering
    \hspace*{-0.25in}
    \includegraphics[width=0.4\linewidth]{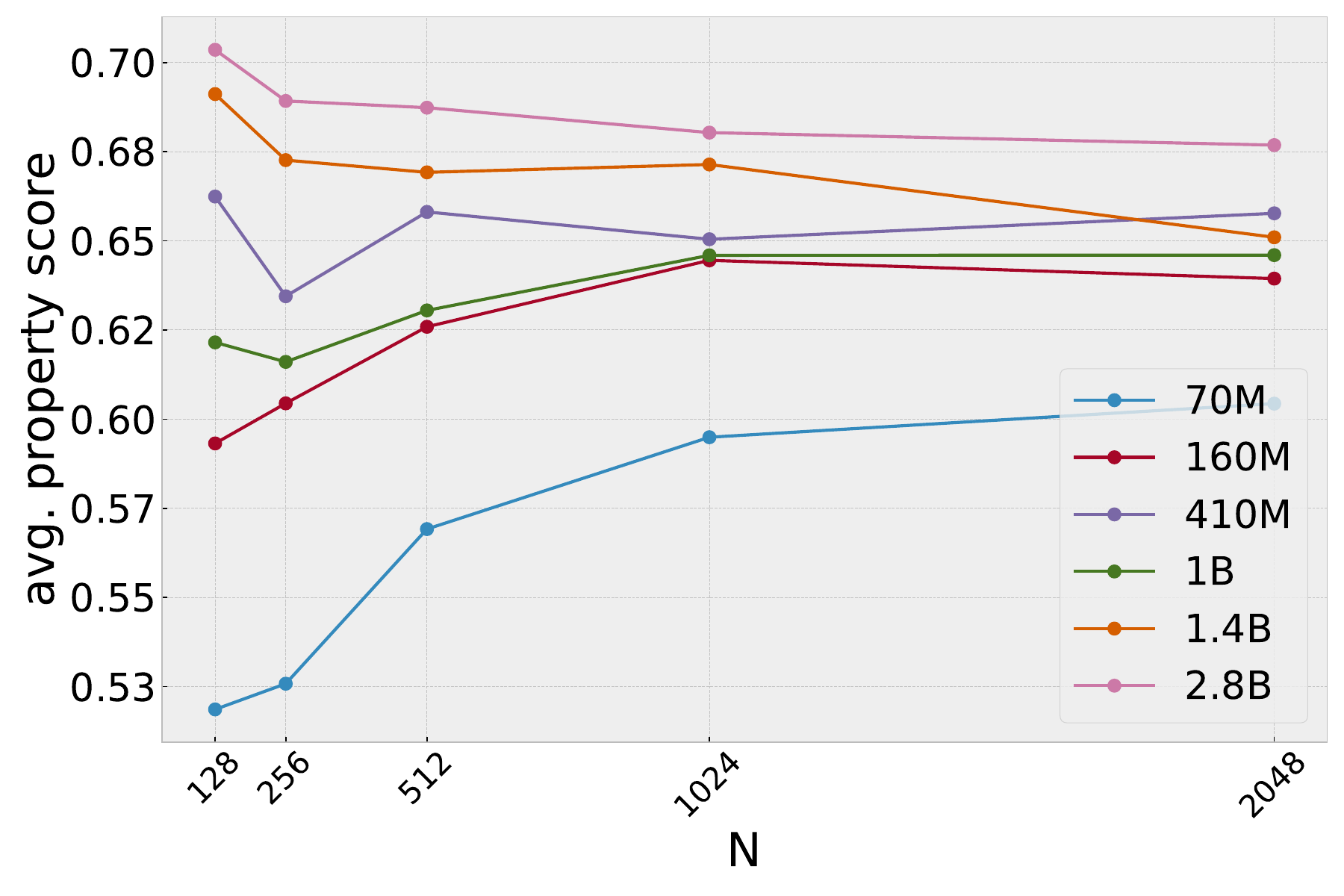}
    \caption{Average task score at increasing $N$ generated samples for different model sizes. We observe diminishing returns past 1B.}
    \label{fig:size_scaling_analysis}
\end{figure}

We analyze how scaling the model size with fixed compute and data affects sample quality. We choose the Pythia scaling suite \cite{biderman2023pythiasuiteanalyzinglarge}, which offers an exhaustive set of model sizes, and we pre-train all the variants up to 2.8B with our dataset. At test time, we measure the average task score for growing sets of molecules to observe quality over number of generations $N$. Figure \ref{fig:size_scaling_analysis} reports comparable performance for all models above 160M of parameters, with diminishing returns past 1B. We choose to fix the default model size to 1B as a tradeoff between model capability and dataset size, in line with \citet{cavanagh2024smileyLlamamodifyinglargelanguage, zholus2024bindgptscalableframework3d}.

\subsection{Analyzing the preference-task correlation}
\label{section:correlation_analysis}

\begin{figure}[H]
    \centering
    \hspace*{-0.25in}
    \includegraphics[width=0.4\linewidth]{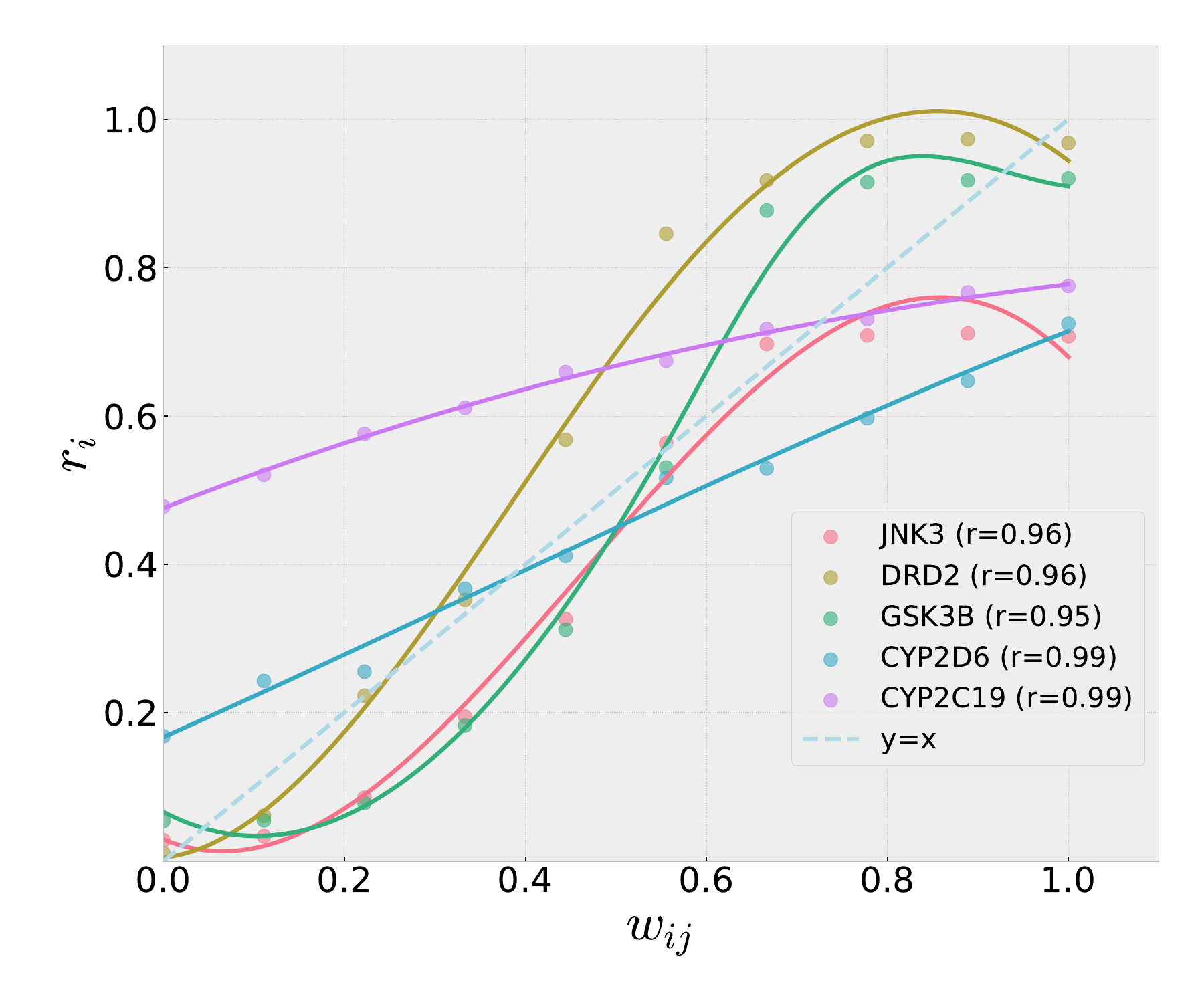}
    \caption{The correlation between task magnitude $w_{ij}$ and $r_i$ is overall positive, with varying value ranges e.g. for \texttt{JNK3} and \texttt{GSK3$\beta$}.}
    \label{fig:correlation_analysis}
\end{figure}

The effectiveness of task arithmetics \citet{ramé2023rewardedsoupsparetooptimalalignment, yadav2023tiesmergingresolvinginterferencemerging} is based on the intuition that the merging coefficients $\textbf{w}_j \in \mathcal{W}$ control the ``magnitude'' of task directions wrt. a base model, $\Delta\theta = \theta_i - \theta_\text{base}$, that is the distribution shift towards a specific objective. In interpolating task directions, we question how directly we can modulate them with defined coefficients. We thus analyze the correlation between preferences $\textbf{w}_j \in \mathcal{W}$  and measured property scores $\{r\}_{i \in \mathbb{R}^{m}}$ in merging pairs of policies. We consider 10 task magnitudes per molecule property $i$, so that $w_{ij} \in [0.1, 1.0] \,,\, \textbf{w}_j = [w_{ij}, 1-w_{ij}]$ for the expert and the base model respectively. A visualization of measured property scores by task magnitude is displayed in Figure \ref{fig:correlation_analysis}. We report an average Pearson correlation coefficient of $0.97$, indicating a direct relationship between $w_{ij}$ and $r_i$. We notice ``saturation points'' varying by property, e.g. \texttt{JNK3}, \texttt{CYP2C19}, approaching values observed at training time. The range of observed rewards not necessarily matches with the provided task magnitudes, e.g. $w_\texttt{CYP2C19} \in [0, 1]$ while $r_\texttt{CYP2C19} \in [0.4, 0.8]$ . We infer that input-output mappings could be derived to better adapt conditioning preferences to each property range, as proposed by \citet{yang2024rewardsincontextmultiobjectivealignmentfoundation}

\pagebreak
\section{Additional experimental results}
\label{appendix:A}

\begin{table}[H]
    \centering
    \scriptsize
    \small
    \begin{tabular}{cccccccccc}
        \toprule
        Method & JNK3 & DRD2 & GSK3$\beta$ & CYP2D6 & CYP2C19 & Uniqueness & Diversity & Average \\
        \midrule
        MORLHF & 0.13 & 0.76 & 0.23 & \textbf{0.89} & 0.81 & 0.94 & \textbf{0.87} & 0.66 \\
        Rewards in Context & 0.55 & \textbf{0.85} & 0.70 & 0.87 & \textbf{0.92} & 0.49 & 0.82 & 0.74 \\
        Rewarded Soups & 0.60 & 0.80 & 0.73 & 0.71 & 0.79 & \textbf{0.95} & 0.84 & 0.77 \\
        \molmoe & \textbf{0.63} & 0.78 & \textbf{0.75} & 0.81 & 0.73 & 0.94 & 0.86 & \textbf{0.78} \\
        \bottomrule
    \end{tabular}
    \caption{Average \textbf{in-distribution task scores} by training method, visualized in Figure \ref{fig:maximization_evaluation}. Overall, \molmoe achieves the highest average score, comparably to RS and RiC. RiC generates a significant fraction of repeated samples.}
    \label{tab:max_points}
\end{table}

\begin{table}[H]
    \centering
    \scriptsize
    \small
    \begin{tabular}{cccccccccc}
        \toprule
        Method & JNK3 & DRD2 & GSK3$\beta$ & CYP2D6 & CYP2C19 & Uniqueness & Diversity & Average \\
        \midrule
        MORLHF & 0.22 & 0.73 & 0.36 & 0.79 & \textbf{0.90} & 0.95 & \textbf{0.87} & 0.69 \\
        Rewards in Context & 0.36 & 0.69 & 0.42 & 0.53 & 0.57 & 0.92 & 0.86 & 0.62 \\
        Rewarded Soups & 0.54 & 0.78 & 0.69 & 0.72 & 0.79 & \textbf{0.96} & 0.86 & 0.76 \\
        \molmoe & \textbf{0.60} & \textbf{0.85} & \textbf{0.71} & \textbf{0.82} & 0.72 & 0.94 & 0.86 & \textbf{0.79} \\
        \bottomrule
    \end{tabular}
    \caption{Average \textbf{out-distribution task scores} by training method, visualized in Figure \ref{fig:maximization_evaluation}. Overall, \molmoe achieves the highest average score, comparably to RS. RiC considerably underperforms the alternative methods, indicating failure in extrapolating to molecules beyond the training distributions.}
    \label{tab:ood_max_points}
\end{table}

\begin{figure}[H]
    \centering
    \includegraphics[width=0.40\textwidth]{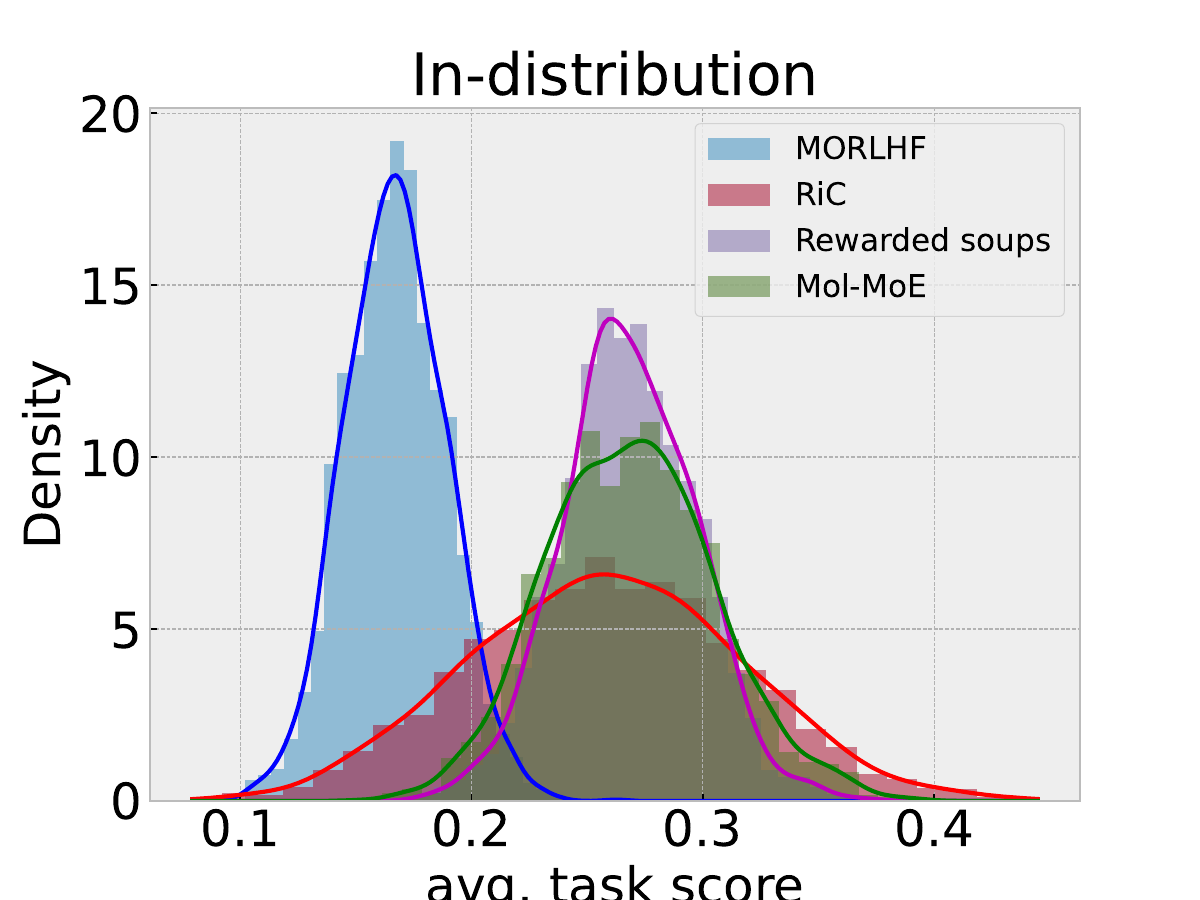}
    \includegraphics[width=0.40\textwidth]{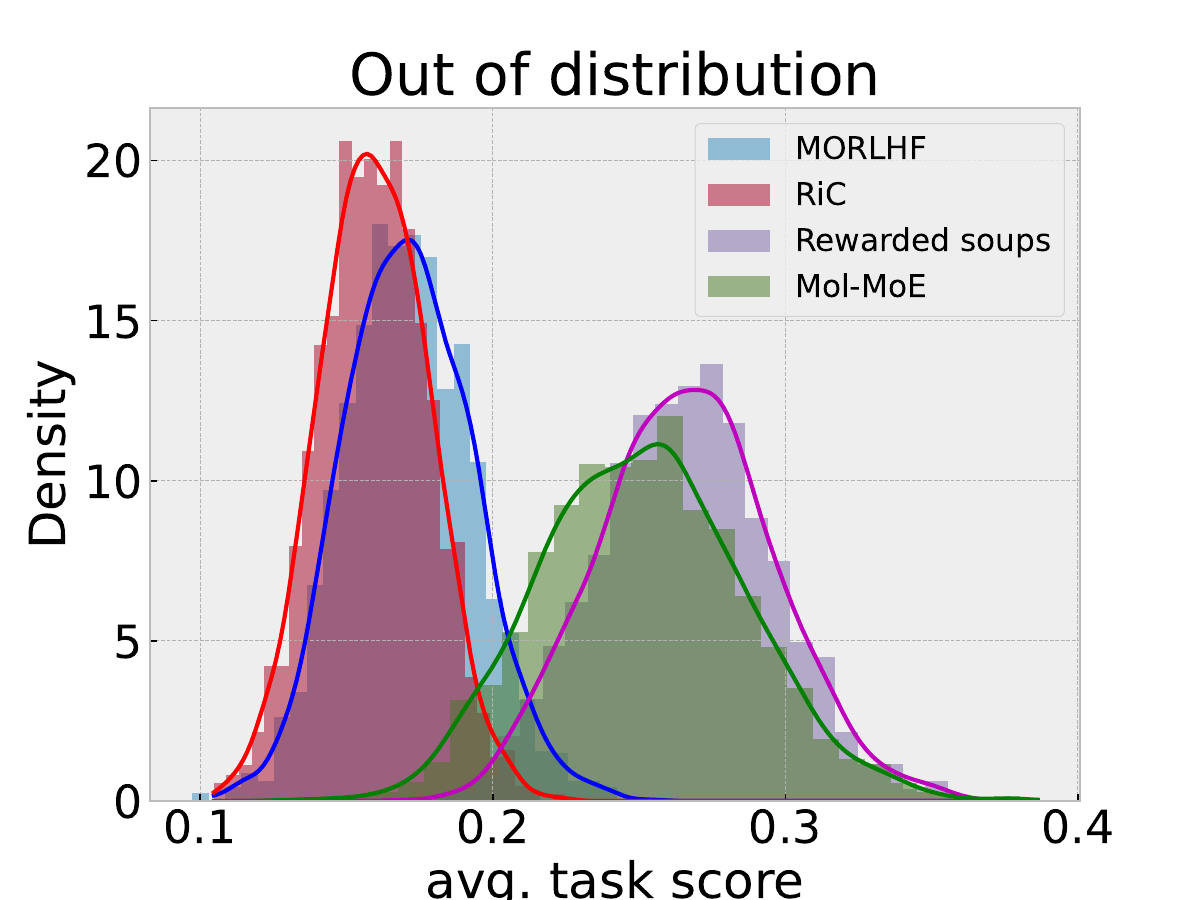}
    \caption{Quality distributions by task score of $2,048$ generated molecules from models trained on the complete dataset (left) or the ablation, held-out best samples (right). \textbf{The distribution from \molmoe is centered on higher scores than most of the alternatives, with a higher density in the right tail especially out of distribution.} Rewards in Context significantly worsens when high quality molecules are removed at training time.}
    \label{fig:distributions}
\end{figure}

\begin{figure}[H]
    \centering
    \hspace*{-0.3in}
    \includegraphics[width=1.0\linewidth]{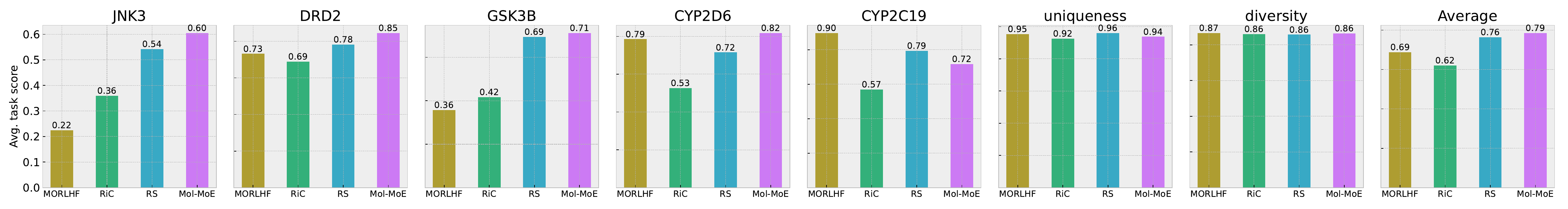}
    \caption{Average property score by tuning method with held-out high quality molecules from the training distribution. \textbf{The distribution of molecules obtained with \molmoe is comparable or superior in sample quality wrt. alternative methods}. The gap in quality between RiC and RS, \molmoe is evident in this setup.}
    \label{fig:ood_task_scores}
\end{figure}

\hfill

\end{document}